\documentclass[letterpaper]{article} % DO NOT CHANGE THIS
\usepackage{aaai2026}  % DO NOT CHANGE THIS
\usepackage{times}  % DO NOT CHANGE THIS
\usepackage{helvet}  % DO NOT CHANGE THIS
\usepackage{courier}  % DO NOT CHANGE THIS
\usepackage[hyphens]{url}  % DO NOT CHANGE THIS
\usepackage{graphicx} % DO NOT CHANGE THIS
\usepackage{natbib}  % DO NOT CHANGE THIS AND DO NOT ADD ANY OPTIONS TO IT
\usepackage{caption} % DO NOT CHANGE THIS AND DO NOT ADD ANY OPTIONS TO IT
\usepackage{algorithm}
\usepackage{algorithmic}

\usepackage{newfloat}
\usepackage{listings}
\DeclareCaptionStyle{ruled}{labelfont=normalfont,labelsep=colon,strut=off} % DO NOT CHANGE THIS
\floatstyle{ruled}
\newfloat{listing}{tb}{lst}{}
\floatname{listing}{Listing}
\newcommand{\trans}{\mathrm{T}}

\usepackage{booktabs}
\usepackage{spreadtab}
\usepackage{amssymb}
\usepackage{amsmath}
\usepackage{bm}
\usepackage{subfigure}
\usepackage{subcaption}
\usepackage{color}
\usepackage{pdfpages}

\title{MeMoSORT: Memory-Assisted Filtering and Motion-Adaptive Association Metric for Multi-Person Tracking}
\author{
    Yingjie Wang, Zhixing Wang, Le Zheng\thanks{Corresponding author.}, Tianxiao Liu, Roujing Li, Xueyao Hu
}
\affiliations{
    Beijing Institute of Technology
}

\usepackage{bibentry}
\begin{document}

\maketitle

\begin{abstract}
Multi-object tracking (MOT) in human-dominant scenarios, which involves continuously tracking multiple people within video sequences, remains a significant challenge in computer vision due to targets' complex motion and severe occlusions. Conventional tracking-by-detection methods are fundamentally limited by their reliance on Kalman filter (KF) and rigid Intersection over Union (IoU)-based association. The motion model in KF often mismatches real-world object dynamics, causing filtering errors, while rigid association struggles under occlusions, leading to identity switches or target loss. To address these issues, we propose MeMoSORT, a simple, online, and real-time MOT tracker with two key innovations. First, the Memory-assisted Kalman filter (MeKF) uses memory-augmented neural networks to compensate for mismatches between assumed and actual object motion. Second, the Motion-adaptive IoU (Mo-IoU) adaptively expands the matching space and incorporates height similarity to reduce the influence of detection errors and association failures, while remaining lightweight. Experiments on DanceTrack and SportsMOT show that MeMoSORT achieves state-of-the-art performance, with HOTA scores of 67.9\% and 82.1\%, respectively.

\end{abstract}

\section{Introduction}

    \begin{figure}[t]
        \centering
        \includegraphics[width=0.95\linewidth]{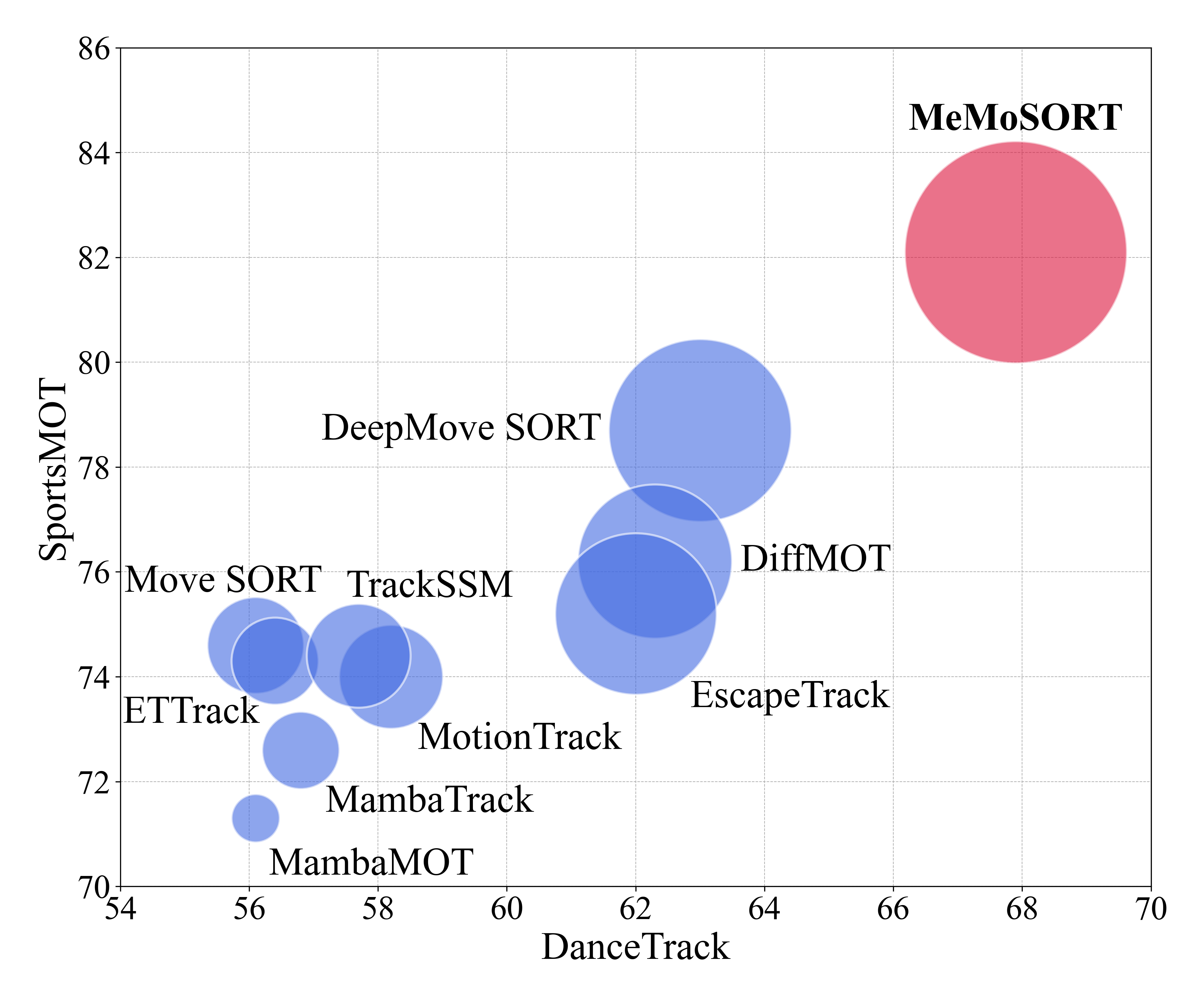} 
        \caption{Performance overview of MeMoSORT against SOTA trackers on the DanceTrack and SportsMOT benchmarks. Each bubble represents a tracker, with its position determined by HOTA scores on the DanceTrack (x-axis) and SportsMOT (y-axis) datasets. The bubble radius is proportional to the average AssA score across both datasets. Our MeMoSORT~(highlighted in red) achieves 67.9\% HOTA, 54.3\% AssA on DanceTrack and 82.1\% HOTA, 75.6\% AssA on SportsMOT, surpassing all other compared trackers significantly.}
        \label{fig: performance}
    \end{figure}

    Multi-object tracking (MOT) is one of the fundamental computer vision tasks that aims at continuously tracking objects in video sequences. 
    It has been widely applied in autonomous driving~\cite{geiger2012kitti, yu2020bdd100k}, video surveillance~\cite{milan2016mot16, dendorfer2020mot20}, sports analysis~\cite{cui2023sportsmot, cioppa2022soccernettracking, sun2022dancetrack}. 

    As the dominant paradigm of MOT, tracking-by-detection (TBD)~\cite{bewley2016sort, zhang2022bytetrack, cao2023ocsort, maggiolino2023deepocsort} addresses this task by decomposing it into three key stages: detector, state estimation (filter), and association. While detection accuracy was historically a primary limiting factor, the advent of high-performance detectors like YOLO~\cite{redmon2016yolo, varghese2024yolov8} has largely addressed this issue. As a result, the performance of modern TBD trackers is now principally constrained by the efficacy of the other two modules: state estimation and association.

    Conventional state estimation and association modules suffer from two key limitations. First, the Kalman filter (KF) \cite{kalman1960kf} assumes linear dynamics and a first-order Markovian process \cite{khodarahmi2023review}, which mismatched with the complex and temporally correlated motion patterns of real-world targets. 
    The mismatch can lead to significant errors in motion prediction and estimation when the actual motion deviates from these assumptions \cite{wang2025transformer}, such as in coordinated or repetitive behaviors (e.g., a dancer consistently spinning after a specific jump). Second, standard association strategies often rely on simplistic Intersection over Union (IoU)~\cite{yu2016iou}, without adapting to the target’s motion patterns. This lack of adaptability can degrade association performance in the presence of complex motion or severe occlusion, potentially resulting in tracking failure.
    
    To address these challenges, we propose MeMoSORT, a simple, online, and real-time MOT framework tailored for complex scenarios. MeMoSORT introduces two key innovations:
    \begin{itemize}
        \item Memory-assisted Kalman Filter (MeKF), which leverages memory-augmented neural networks to compensate for the gap between assumed and actual motion patterns.
        \item Motion-adaptive IoU (Mo-IoU), which adaptively expands the matching space and incorporates height similarity to reduce association errors, while remaining computationally efficient.
    \end{itemize}

    Extensive experiments demonstrate that MeMoSORT achieves state-of-the-art (SOTA) performance on challenging benchmarks, reaching HOTA scores of 67.9\% on DanceTrack and 82.1\% on SportsMOT, significantly outperforming existing methods across multiple metrics (see Figure~\ref{fig: performance}).

\section{Related Works}

    \subsection{Methods for Motion Estimation}
    KF is the widely used motion estimator in early TBD trackers. Subsequent methods such as OC-SORT~\cite{cao2023ocsort} introduced improvements to handle occlusions, but could not overcome the fundamental limitations of the linear, first-order Markovian motion model in scenarios with complex, non-linear dynamics.

    To address this, one line of research replaces the KF entirely with data-driven neural networks. For example, DiffMOT~\cite{lv2024diffmot} employs a diffusion model as non-linear motion prediction, while Mamba-based trackers~\cite{xiao2024mambatrack, khanna2025sportmamba} utilize State-Space Models to capture complex motion. However, a key challenge for these pure predictors is the lack of a principled filtering step; they often update a track's state directly with the noisy detector measurement, which can degrade trajectory quality.
    
    Another direction~\cite{li2024srtrack, adzemovic2025movesort} involves hybrid approaches that embed deep learning technique within the classic filter structure. These methods combine the expressiveness of neural networks with the stability of the prediction-update cycle. A potential drawback, however, is that fully replacing KF's simple, physics-based prior with a complex, data-driven model makes performance entirely contingent on the training data, potentially harming robustness and generalization.

    \subsection{Association between Detection and Prediction} 

    Mainstream association methods within the TBD paradigm are typically based on two principles: spatial consistency and appearance similarity. The former is primarily addressed by IoU and its variants, while the latter relies mainly on ReID based methods. In practice, these two approaches are often combined into a final association cost, typically through a weighted sum.

    IoU-based methods use IoU as spatial association metric, higher IoU between boxes across frames represents higher probability of the same targets. 
    % C-BIoU~\cite{yang2023cbiou} and Deep-EIoU~\cite{huang2024deepeiou} extends the scale of box to mitigate prediction error.
    % HybridSORT~\cite{yang2024hybridsort} incorporate height similarity into IoU calculation.
    % SportMamba intorduce Height-Adaptive EIoU (HA-EIoU), which is the joint format of EIoU and HMIoU. 
    % However, the performance of above types of IoU critically depends on the parameter controlling the box expansion scale, limiting their applicability in online systems. 
    Recent studies modified IoU by expanding the scale of the box~\cite{yang2023cbiou, huang2024deepeiou}, incorporating height similarity~\cite{yang2024hybridsort} or considering both. 
    However, the performance of above types of IoU with fixed parameters critically depends on manual setting, limiting their applicability in online systems. 
    Existing dynamic parameter methods either use multiple association stages with several fixed parameter~\cite{huang2024deepeiou} or focus on temporal information of the trajectory~\cite{stanojevic2024boosttrack++}, lacking adaptivity according to target's motion characteristics.

    ReID-based methods uses an additional neural networks to extract feature to represent the visual appearance of target, considering shorter distance between feature across frames leads to same target.
    The majority of ReID based methods~\cite{wojke2017deepsort, aharon2022botsort, du2023strongsort} use convolution neural network to extract appearance feature and apply cosine distance as measurement. 
    ReID-based methods are less effective in distinguishing targets with similar appearance or under occlusion.

\section{Methodology}

    \subsection{Preliminaries: Tracking by Detection}
    The TBD paradigm is a prevalent approach in MOT. Unlike monolithic end-to-end methods, TBD frameworks decouple the tracking problem into three distinct stages, as illustrated in Figure~\ref{fig: framework-tbd}: object detection, association, and filtering. 
    
    The first step involves an object detector, such as the popular YOLO model, generates a set of candidate boxes for each frame $t$. A detection is typically represented as a vector $\widetilde{\bm{b}}_{t} = [\widetilde{x}_t, \widetilde{y}_t, \widetilde{w}_t, \widetilde{h}_t]^\trans$, defining the center coordinates, width, and height of the box. It is generated by the linear measurement matrix $\mathbf{H}$ from the target's state $\bm{b}_t$, namely,
    \begin{align}
        \widetilde{\bm{b}}_{t} = \mathbf{H} \bm{b}_t + \bm{v}_t, \label{eq:linear_meas_model}
    \end{align}
    where $\bm{v}_t$ is the measurement noise, it is generally assumed to follow an independent Gaussian distribution with a mean of $\mathbf{0}$ and a covariance matrix $\mathbf{R}_t$.
    
    The output detections, which are prone to false alarms and misses from occlusion, are linked across frames via data association to form trajectories. This task is formulated as a bipartite matching problem between existing tracks and current detections and is optimally solved using the Hungarian algorithm. The matching cost is determined by the IoU and ReID. Specifically, IoU measures the spatial overlap between a detection $\widetilde{\bm{b}}_{t}$ and a track's predicted state $\hat{\bm{b}}'_{t}$. And ReID involves masking the object within the detection box, encoding its appearance, and then measuring similarity using cosine distance.

    After association, the KF recursively estimates the target's state via a prediction-update cycle. The prediction is based on a linear, first-order Markovian motion model:
    \begin{align}
        \bm{b}_t = \mathbf{F} \bm{b}_{t-1} + \bm{w}_t, \label{eq:markovian_motion_model}
    \end{align}
    where $\textbf{F}$ is the linear state transition matrix (e.g. constant velocity model). And $\bm{w_t}$ is the process noise, it is generally assumed to follow an independent Gaussian distribution with a mean of $\mathbf{0}$ and a covariance matrix $\mathbf{Q}_t$. In the update step, this prediction is refined by incorporating the newly associated detection.   
    
    However, this prevalent pipeline suffers from two critical limitations. First, the state estimation relies on an underlying linear, first-order Markovian motion model is often an oversimplification of real-world dynamics. This prevents the KF from handling complex, non-linear paths or re-identifying targets after prolonged occlusions. Second, the association cost, based mainly on IoU, is unreliable during occlusion as the boxes is mixed to a mess. To this end, our work introduces a deep learning aided filter that leverages temporal memory to model complex dynamics and a robust association metric resilient to occlusion.

    \subsection{Framework of the Proposed MeMoSORT}
    
    \begin{figure}[t]
        \centering
            \subfigure[Framework of Tracking-by-Detection]{
                \label{fig: framework-tbd}
                \includegraphics[width=0.95\linewidth]{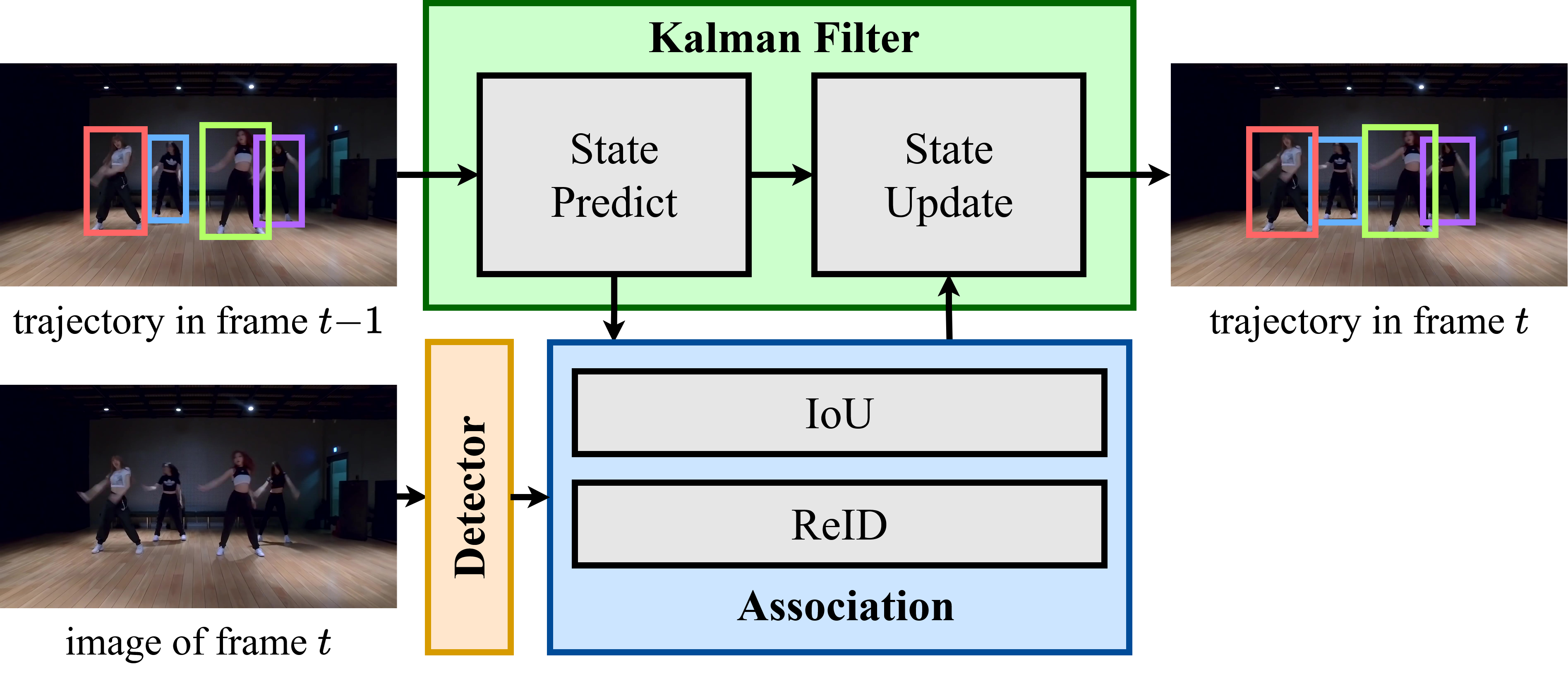}
            }
            \quad
            \subfigure[Framework of MeMoSORT]{
                \label{fig: framework-our}
                \includegraphics[width=1\linewidth]{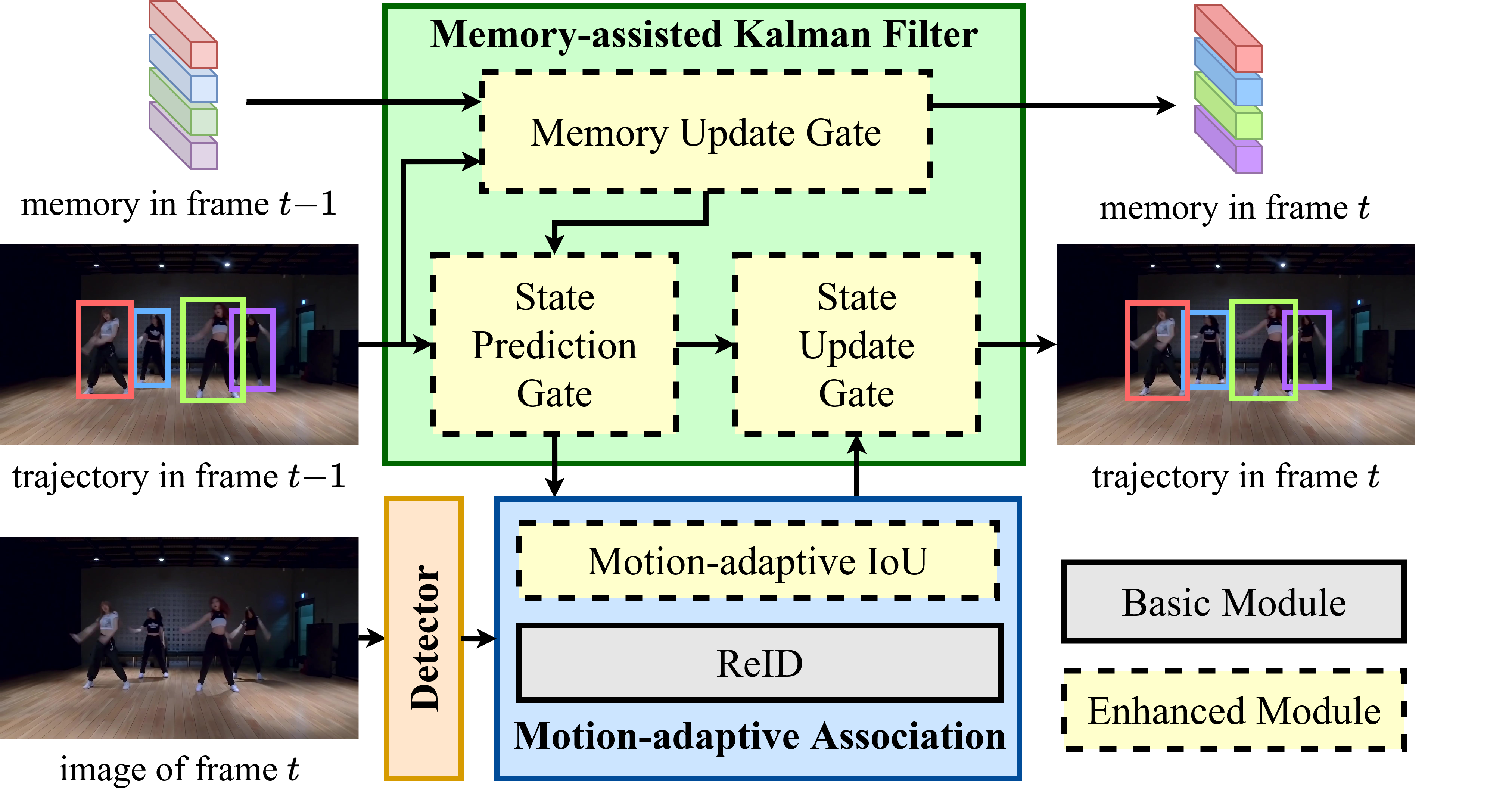}
            }
        \caption{
        Comparison between (a) the conventional Tracking-by-Detection framework and (b) our proposed MeMoSORT framework. MeMoSORT introduces two key enhancements: it leverages a memory mechanism to guide motion estimation for more accurate state prediction and update, and it applies a Motion-adaptive IoU to achieve robust data association. Together, these improvements enhance tracking robustness in complex scenarios.}
    \end{figure}

    The framework of our proposed MeMoSORT is illustrated in Figure~\ref{fig: framework-our}, with the following three key components.

        \subsubsection{Detection} In line with the conventional TBD paradigm, MeMoSORT leverages the YOLOX~\cite{ge2021yolox} to perform the initial detection task, generating a set of candidate bounding boxes for all potential targets within each frame.

        \subsubsection{Association} We introduce an association pipeline inspired by Deep OC-SORT \cite{deepocsort}. This pipeline incorporates our novel Mo-IoU, a metric that refines conventional IoU by adaptively expanding the boxes and considering height similarity based on the target's motion characteristics. Within this pipeline, detections are initially stratified by their confidence scores. High-scoring detections are matched using a combined Mo-IoU and ReID cost via the Hungarian algorithm, while low-scoring detections are matched using a standard IoU cost.    
        
        \subsubsection{Filter} We propose the MeKF, a variant of the standard KF inspired by literature \cite{shiyanegbrnn} that leverages memory to aid in state estimation. The MeKF consists of three gated modules: a Memory Update Gate (MUG) to maintain a historical representation, a State Prediction Gate (SPG) to correct the motion prediction using memory, and a State Update Gate (SUG) to refine the state based on the associated detection.        

    \subsection{Memory-Assisted Kalman Filter}

    To address the limitations of the first-order Markovian assumption in the KF (Eq. \ref{eq:markovian_motion_model}), we introduce a non-Markovian motion formulation capable of modeling the complex dynamics inherent in real-world targets:
    \begin{align}
        \bm{b}_t = f_t ( \bm{b}_{t-1}, \bm{b}_{t-2}, ..., \bm{b}_1) + \bm{w}_t, \label{eq:non-markovian_motion_model}
    \end{align}
    where, $f_t(\bm{\cdot})$ is a non-linear transition function. Unlike the first-order Markovian state transition matrix $\mathbf{F}$ in Eq. \ref{eq:markovian_motion_model}, $f_t(\bm{\cdot})$ explicitly conditions the state prediction on the full trajectory history, thus enabling the modeling of long-term dependencies.

    As an explicit analytical form for $f_t(\bm{\cdot})$ is intractable, we simplify the problem by introducing a memory mechanism. Specifically, we construct a memory representation $\bm{m}_t$ and update it recursively using a memory update function $\phi_t (\bm{\cdot})$, namely,
    \begin{align}
        \bm{m}_t &= \phi_t (\bm{b}_{t-1}, \bm{b}_{t-2}, ..., \bm{b}_{1}), \nonumber \\
                 &\approx \phi_t (\bm{b}_{t-1}, \bm{m}_{t-1}). \label{eq:memory_update}
    \end{align}
    
    Subsequently, a state compensation function $\psi_t (\bm{\cdot})$ generates a compensation $\bm{\Delta}^\mathbb{F}_t$ for the linear Markovian state transition process, namely, $\bm{\Delta}^\mathbb{F}_t = \psi_t (\bm{m}_t)$. Thereby approximating the non-markovian motion model in Eq.~\ref{eq:non-markovian_motion_model} by a first-order Markovian manner. This process can be formulated as:
    \begin{gather}
        \bm{b}_t \approx \mathbf{F} \bm{b}_{t-1} + \bm{\Delta}^\mathbb{F}_t + \bm{w}_t. \label{eq:approx_non-Markovian}
    \end{gather}
    
    Furthermore, the linear measurement matrix $\mathbf{H}$ defined in Eq. \ref{eq:linear_meas_model}, often fails to represent the true observation process. To address this discrepancy, a similar approximation can be made, i.e.,
    \begin{gather}
        \widetilde{\bm{b}}_{t} \approx \mathbf{H} \bm{b}_t + \bm{\Delta}^\mathbb{H}_t + \bm{v}_t, \label{eq:approx_true_meas}
    \end{gather}
    where the compensation term $\bm{\Delta}^\mathbb{H}_t$ is generated by $\widetilde{\bm{b}}_{t}$ through function $\varphi_t (\bm{\cdot})$, namely, $\bm{\Delta}^\mathbb{H}_t {=} \varphi_t (\widetilde{\bm{b}}_{t})$.

\begin{figure}[t]
        \centering
        \includegraphics[width=0.95\linewidth]{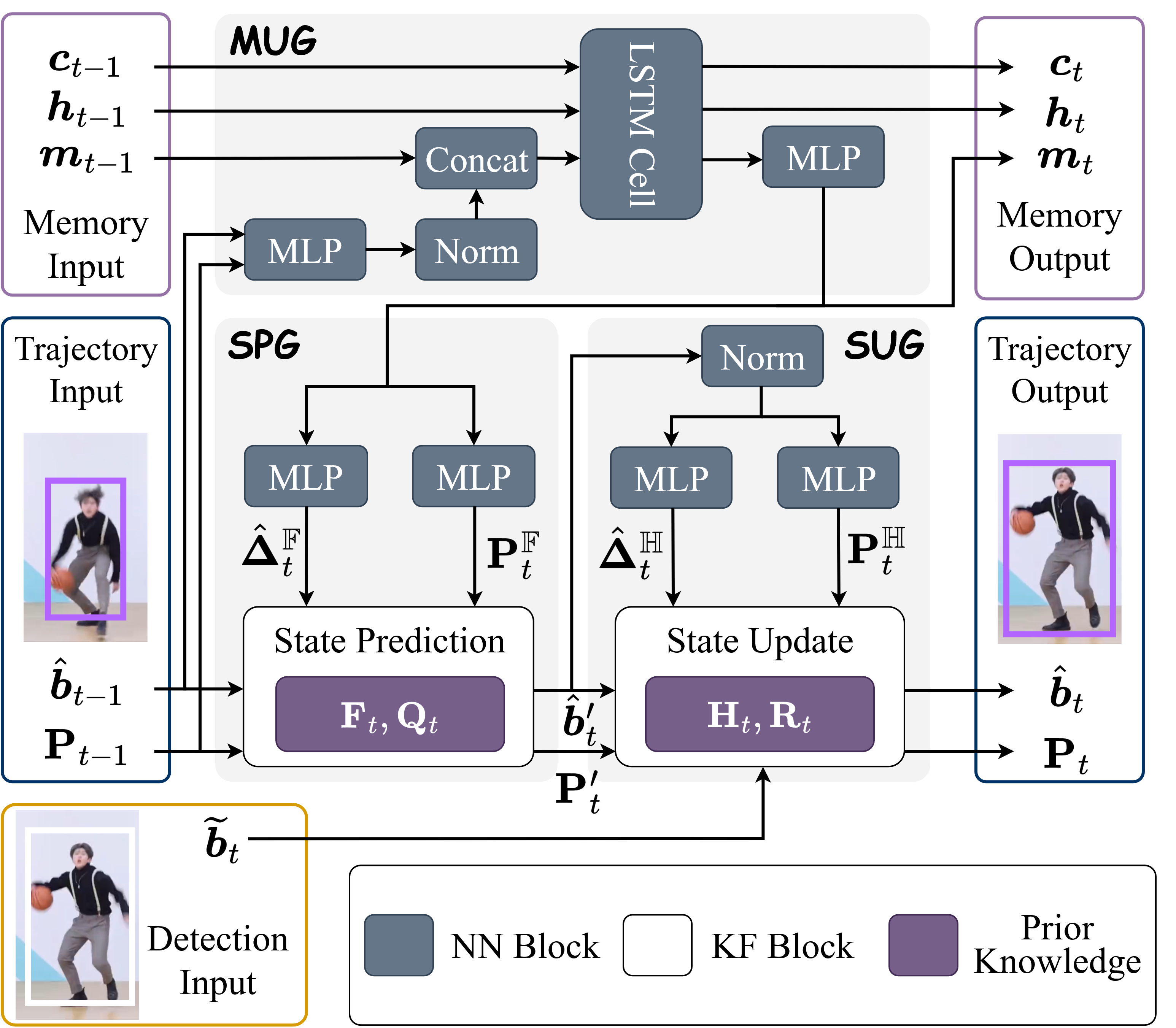}
        \caption{Framework of proposed MeKF. In both SPG and SUG, the NN blocks assist the physical motion model by generating compensation for its errors, based on memory and detection respectively. This approach robustly ensures the stability of the state estimation; even if the NN fails, the underlying physical model can still provide a baseline prediction as a failsafe.}
        \label{fig:framework_mekf}
    \end{figure}

    However, the memory update function $\phi_t (\bm{\cdot})$, state compensation function $\psi_t (\bm{\cdot})$, and measurement compensation function $\varphi_t (\bm{\cdot})$ are difficult to model with explicit analytical forms. Such that we employ neural network (NN) technique to fit these complex, non-linear functions. By integrating these learned modules with the foundational principles of Eqs. \ref{eq:memory_update} - \ref{eq:approx_true_meas}, we construct a data-driven Bayesian filter: the MeKF, as shown in Figure \ref{fig:framework_mekf}. 

        \subsubsection{Memory Update Gate} The memory update process in Eq.~\ref{eq:memory_update} is formally analogous to that of a Recurrent Neural Network. We therefore implement the update function $\phi_t(\bm{\cdot})$ using the Long Short-Term Memory (LSTM) network. The LSTM is trained to distill and update the memory from the historical trajectory sequence, with the specific update process detailed as follows:
        \begin{gather}
            \bm{m}_t = \mathcal{F}_{\mathrm{LSTM}} (\bm{c}_{t-1}, \bm{h}_{t-1}, \bm{m}_{t-1}), \label{eq:MUG}
        \end{gather}
        where $\mathcal{F}_{\mathrm{LSTM}}(\bm{\cdot})$ denotes the mapping function of the MUG, implemented by the LSTM network. And $\bm{c}_{t-1}$ and $\bm{h}_{t-1}$ are the cell state and hidden state of the LSTM, respectively.
    
        \subsubsection{State Prediction Gate} In contrast to MoveSORT and DiffMOT, which directly utilize NN to predict the target's state, the SPG compensates for the error between the physical motion model and the true physical process. While reducing the amount of parameters, the SPG leverages a prior model to guarantee the error lower bound of the MeKF, which is defined as follows:
        \begin{gather}
            \hat{\bm{b}}'_{t} = \mathbf{F} \hat{\bm{b}}_{t-1} + \hat{\bm{\Delta}}^\mathbb{F}_t \label{eq:SPG_f}, \\
            \mathbf{P}'_{t} = \mathbf{F} \mathbf{P}_{t-1} \mathbf{F}^\trans + \mathbf{P}^\mathbb{F}_t + \mathbf{Q}_t, \label{eq:SPG_s}
        \end{gather}
        where $\hat{\bm{b}}'_{t}$ and $\mathbf{P}'_{t}$ represent the state prediction and the error covariance prediction, respectively. Here, $\hat{\bm{\Delta}}^\mathbb{F}_t {=} \mathcal{F}_{\mathrm{MLP}}^1 (\bm{m}_t)$ and $\mathbf{P}^\mathbb{F}_t {=} \mathcal{F}_{\mathrm{MLP}}^2 (\bm{m}_t) (\mathcal{F}_{\mathrm{MLP}}^2 (\bm{m}_t))^\trans$ are the exception and covariance compensation generated by distinct multilayer perceptrons (MLP) with unshared parameters.
    
        \subsubsection{State Update Gate} Similarly, the SUG utilizes distinct MLPs to generate corresponding compensation terms and is naturally embedded within the state update process, namely,
        \begin{gather}
            \mathbf{K}_t = \mathbf{P}'_t \mathbf{H}^\trans (\mathbf{H} \mathbf{P}'_t \mathbf{H}^\trans + \mathbf{R}_t + \mathbf{P}^\mathbb{H}_t)^{-1}, \label{eq:kalman_gain} \\
            \hat{\bm{b}}_t = \hat{\bm{b}}'_{t} + \mathbf{K}_t (\widetilde{\bm{b}}_t - \mathbf{H} \hat{\bm{b}}'_{t} - \hat{\bm{\Delta}}^\mathbb{H}_t) ,\label{eq:SUG_f} \\
            \mathbf{P}_t = (\mathbf{I} - \mathbf{K}_t \mathbf{H}) \mathbf{P}'_t \label{eq:SUG_s},
        \end{gather} 
        where $\hat{\bm{b}}_t$ and $\mathbf{P}_{t}$ are the state update and the error covariance update, respectively. Here, $\hat{\bm{\Delta}}^\mathbb{H}_t = \mathcal{F}_{\mathrm{MLP}}^3 (\hat{\bm{b}}'_{t})$, $\mathbf{P}^\mathbb{H}_t = \mathcal{F}_{\mathrm{MLP}}^4 (\hat{\bm{b}}'_{t}) (\mathcal{F}_{\mathrm{MLP}}^4 (\hat{\bm{b}}'_{t}))^\trans$ and $\mathbf{K}_t$ is the Kalman gain. All of the aforementioned gates are designed based on Bayesian principles similar as KF and are supported by theoretical derivations, which are detailed in Appendix A.

    % \subsection{Motion-Adaptive IoU for Association}
    \subsection{Motion-Adaptive Association}

    To achieve robust association in complex scenarios such as occlusion, we introduce the Motion-adaptive IoU (Mo-IoU). It is defined as a multiplicative fusion of two novel components designed to address distinct challenges:
    \begin{align}
        & \mathrm{Mo\mbox{-}IoU} (\hat{\bm{b}}'_{t}, \widetilde{\bm{b}}_t, p_t, q_t) \nonumber \\
        = & \mathrm{EIoU}(\hat{\bm{b}}'_{t}, \widetilde{\bm{b}}_t, p_t) \times \mathrm{HIoU}(\hat{\bm{b}}'_{t}, \widetilde{\bm{b}}_t, q_t), \label{eq:Mo-IoU}
    \end{align}
    where Expansion IoU (EIoU) adaptively expands matching space to maintain tracking against position errors of the detector, and Height IoU (HIoU) emphasizes height similarity to distinguish occluded targets. Their product imposes a strong joint constraint, and the parameters $p_t$ and $q_t$ are adaptively set by our Motion-Adaptive Technique (MAT).
        
        \subsubsection{Expansion IoU}
        To strengthen tracking robustness against position errors of the detector, we introduce EIoU. It operates by scaling the width and height of the original box pair before computing their IoU, an operation equivalent to:
        \begin{align}
            \mathrm{EIoU}(\hat{\bm{b}}'_{t}, \widetilde{\bm{b}}_t, p_t) = \mathrm{IoU}(\hat{\bm{e}}'_{t}, \widetilde{\bm{e}}_t), \label{eq:EIoU}
        \end{align}
        where $\hat{\bm{e}}'_{t} {=} [\hat{x}'_{t}, \hat{y}'_{t}, (2 p_t {+} 1) \hat{w}'_{t}, (2 p_t {+} 1) \hat{h}'_{t}]^\trans$ and $\widetilde{\bm{e}}_t {=} [\widetilde{x}_t, \widetilde{y}_t, \\ (2 p_t {+} 1) \widetilde{w}_t, (2 p_t {+} 1) \widetilde{h}_t]^\trans$ are the expansion boxes. The scaling factor $(2 p_t {+} 1)$ is chosen as it allows EIoU to gracefully degenerate to the standard IoU when $p_t{=}0$, while providing a symmetric expansion. This increased size ensures a greater chance of overlap, preventing association failures even in the presence of significant position errors.
        
        \subsubsection{Height IoU}
        In severe occlusion scenarios, standard IoU often fails, yet a target's height can remain a highly distinguishable feature. To leverage this observation, we define HIoU as:
        \begin{equation}
            \mathrm{HIoU}(\hat{\bm{b}}'_{t}, \widetilde{\bm{b}}_t, q_t) 
            = \left( \frac{l_t}{\hat{h}'_{t} + \widetilde{h}_t - l_t} \right) ^ {q_t},
        \label{eq:HIoU}
        \end{equation}
        where $l_t$ denotes the intersection height of $\hat{\bm{b}}'_{t}$ and $\widetilde{\bm{b}}_t$, and the exponent $q_t$ adaptively controls the emphasis placed on this height similarity. The base of this formula is geometrically equivalent to a 1D-IoU on the vertical axis, robustly measuring the boxes' vertical alignment.
        
        \subsubsection{Motion-Adaptive Technique}
        
        To improve the generalization of Mo-IoU in diverse scenarios, a novel Motion-Adaptive Technique (MAT) is proposed to adaptively adjust expansion parameters $p_t$ and $q_t$ based on the target's motion characteristics, which is formulated as:
        \begin{align}
            p_t = &
            \begin{cases}
                M_{\text{slow}} \quad&\text{if } \dot{c}_{t-1} \le \Theta_{\text{center}}, \\
                M_{\text{fast}} \quad&\text{otherwise,}
            \end{cases} \label{eq:adaptive_p_t} \\
            q_t = &
            \begin{cases}
                N_{\text{slow}} \quad&\text{if } \dot{l}_{t-1} \le \Theta_{\text{height}}, \\
                N_{\text{fast}} \quad&\text{otherwise,}
            \end{cases} \label{eq:adaptive_q_t}
        \end{align}
        where $\dot{c}_{t-1} {=} \sqrt{ (\dot{x}_{t-1}/w_{t-1})^2 {+} (\dot{y}_{t-1}/h_{t-1})^2 }$ is the normalized speed of box center, and $\dot{l}_{t-1} {=} \dot{h}_{t-1}/h_{t-1}$ is the normalized speed of box height. The terms $\Theta_{\text{center}}$ and $\Theta_{\text{height}}$ are predefined thresholds for these two speed. As a scale independent variation, the normalized speed is a suitable quantitative description of the target's motion characteristics.

        The parameter $p_t$ compensates for the motion model's prediction error. Since high-speed motion often leads to larger errors, a larger expansion factor ($p_t {=} M_{\text{fast}}$) is used to provide greater spatial tolerance, and vice versa. In contrast, the parameter $q_t$ adapts to the reliability of height as a feature: a rapidly changing, less reliable height warrants a small expansion factor ($q_t {=} N_{\text{fast}}$), and vice versa. Besides, these parameters are quantized into discrete levels. This approach avoids the computational overhead of fractional exponents, ensuring efficiency in real-time systems while maintaining robust tracking performance.

        % As a new adaptive association metric, Mo-IoU adaptively adapts to different motion characteristics in real-time systems by setting proper parameters controlling expansion scale and height importance automatically. The quantization strategy ensures these adaptations incur minimal computational overhead, making Mo-IoU particularly suitable for online tracking systems where both accuracy and efficiency are critical.

        % To create a scale-invariant measure of motion as the basic of MAT, we define the standardize the speed of box $\hat{\bm{b}}'_{t-1}$ by normalizing velocity component $[\dot{x}_{t-1}, \dot{y}_{t-1}, \dot{w}_{t-1}, \dot{h}_{t-1}]^\top = \hat{\bm{b}}_{t-1} - \hat{\bm{b}}_{t-2}$ with the box's width and height, formulate as: 
        % \begin{align}
        %     \dot{c}_{t-1} =& \sqrt{ {\left( \frac{\dot{x}_{t-1}}{w_{t-1}} \right)}^2 + {\left( \frac{\dot{y}_{t-1}}{h_{t-1}} \right)}^2 },
        %     \label{eq:vc}
        %     \\
        %     \dot{l}_{t-1} =& \frac{\dot{h}_{t-1}}{h_{t-1}},
        %     \label{eq:vh}
        % \end{align}
        % where $\dot{c}_{t-1}$ and $\dot{l}_{t-1}$ represents the standardized speed of box center and box height.

\section{Experiments}

    \subsection{Settings}

        \subsubsection{Datasets}
        We conducted the main experiments on DanceTrack and SportsMOT datasets known for their diverse and rapid movements and indistinguishable appearances, in which the performance of ReID module is highly limited, requiring accurate motion capability. 
        DanceTrack is currently one of the most challenging benchmarks in the MOT field, characterized by frequent and severe occlusion as well as highly similar appearance, which requires a significant demand on the trackers’ motion capacity to robustly handle long term identity consistency. 
        SportsMOT introduces sports video sequences with fast, variable-speed motion and extensive camera motion, thereby demanding more robustness in motion model and association. 

        \subsubsection{Metrics}
        We utilize Higher Order Metric~\cite{luiten2021hota} (HOTA, AssA, DetA), IDF1~\cite{ristani2016idf1}, and CLEAR metrics~\cite{bernardin2008clearmot} (MOTA) as our evaluation metrics. 
        Among various metrics, HOTA is the core benchmark that holistically balances association consistency and positional precision.
        Complementing this, IDF1 and AssA specifically measure association quality and identity preservation, while DetA and MOTA primarily evaluate motion estimation accuracy. 
        Additionally, computational efficiency is quantified through frames per second (FPS) to evaluate real-time processing capability.

        \subsubsection{Implementation Details}
        For the training of our proposed MeKF, we utilize AdamW optimizer with learning rate set to $10^{-4}$.
        For our detector, we fine-tune the COCO-pretrained YOLOX model on CrowdHuman~\cite{shao2018crowdhuman} and the target dataset, similar to SportsMOT. 
        In the association stage, the confidence threshold of high-score and low-score matching are set to 0.6 and 0.1. 
        For the Mo-IoU component, the height modulation parameter are set to $M_{\text{slow}} {=} 2$, with $M_{\text{fast}} {=} M_{\text{slow}} {-} 1$, with expansion scaling parameters uses $N_{\text{slow}} {=} 0.5$ and $N_{\text{fast}} {=} N_{\text{slow}} {+} 0.1$. Velocity thresholds $\Theta_{\text{center}}$ and $\Theta_{\text{height}}$ are determined by the 50th and 70th percentile of the normalized velocity distribution from training set (i.e. 0.0406 and 0.0090 for DanceTrack, 0.1172 and 0.0062 for SportsMOT).
        All other unspecified hyper-parameters and ReID model are kept consistent with those in Deep OC-SORT.

    \subsection{Benchmark Results}
    We evaluates MeMoSORT~against state-of-the-art trackers on DanceTrack and SportsMOT benchmarks. 
    % Empirical results demonstrate consistent superiority across both datasets, achieving performance gains with marginal computational overhead while preserving core "SORT" attributes: Simple, Online and Real-Time. 
    For those methods which proposed several kinds of model, we always put the one with the highest HOTA in the table. 

        \subsubsection{DanceTrack}
        As depicted in Table \ref{tab:result-dance}, MeMoSORT~accomplishes the best results among all methods with 67.9\% HOTA, 54.3\% AssA, 85.0\% DetA and 68.0\% IDF1, and remains competitive performance on MOTA. 
        Compared with the previous SOTA tracker TrackTrack~\cite{shim2025tracktrack}, MeMoSORT~outperforms it in HOTA by 1.4\%. 
        The superior result of AssA (54.3\%) underscores the effectiveness of our proposed Mo-IoU in maintaining correct identities through complex interactions and long-term occlusions. 
        Furthermore, the exceptional result of DetA (85.0\%) highlights the capability of our MeKF network to accurately model the targets' abrupt and non-linear movements. 
        It is worth noting that many recent top-performing methods, such as DeepMove SORT and DiffMOT, rely on large, data-driven models like Transformer and Diffusion. 
        In contrast, MeMoSORT~adopts a simple yet powerful approach. 
        By effectively leveraging prior motion information, it achieves superior performance without such architectural complexity, demonstrating a more elegant solution.

        \begin{table}[h]
        \centering
        \setlength{\tabcolsep}{1mm}
        %\resizebox{.95\columnwidth}{!}{
        \begin{tabular}{l|c c c c c}
            \toprule
            % Tracker & HOTA $\uparrow$ & IDF1 $\uparrow$ & MOTA $\uparrow$ & DetA $\uparrow$ & AssA $\uparrow$ \\
            Tracker & HOTA & AssA & IDF1 & DetA & MOTA \\ 
            \midrule
                ByteTrack & 47.7  & 32.1  & 53.9  & 71.0  & 89.6  \\ 
                OC-SORT & 55.1  & 40.4  & 54.9  & 80.4  & 92.2  \\ 
                MoveSORT & 56.1  & 38.7  & 56.0  & 81.6  & 91.8  \\ 
                MambaMOT & 56.1  & 39.0  & 54.9  & 80.8  & 90.3  \\ 
                ET-Track & 56.4  & 39.1  & 57.5  & 81.7  & 92.2  \\ 
                MambaTrack & 56.8  & 39.8  & 57.8  & 80.1  & 90.1  \\ 
                Track SSM & 57.7  & 41.0  & 57.5  & 81.5  & 92.2  \\ 
                MotionTrack & 58.2  & 41.7  & 58.6  & 81.4  & 91.3  \\ 
                C-BIoU & 60.6  & 45.4  & 61.6  & 81.3  & 91.6  \\ 
                Deep OC-SORT & 61.3  & 45.8  & 61.5  & 82.2  & 92.3  \\ 
                CMTrack & 61.8  & 46.4  & 63.3  & - & 92.5  \\ 
                EscapeTrack & 62.0  & 48.6  & 66.4  & - & 89.5  \\ 
                C-TWiX & 62.1  & 47.2  & 63.6  & 81.8  & 91.4  \\ 
                Hybrid-SORT & 62.2  & - & 63.0  & - & 91.6  \\ 
                DiffMOT & 62.3  & 47.2  & 63.0  & \underline{82.5}  & 92.8  \\ 
                DM SORT & 63.0  & 48.6  & 65.0  & 82.0  & 92.6  \\ 
                UCMCTrack & 63.6  & 51.3  & 65.0  & - & 88.9  \\ 
                Hybrid-SORT & 65.7  & - & 67.4  & - & 91.8  \\ 
                TrackTrack & \underline{66.5}  & \underline{52.9}  & \underline{67.8}  & - & \textbf{93.6}  \\ 
                \textbf{MeMoSORT} & \textbf{67.9}  & \textbf{54.3}  & \textbf{68.0}  & \textbf{85.0}  & \underline{93.4}  \\  
            \bottomrule
        \end{tabular}
        \caption{Performance comparison on the DanceTrack test set. For all metrics, higher values indicate better performance. The best and second-best results are highlighted in \textbf{bold} and with an \underline{underline}, respectively. The result of ByteTrack is cited from the original DanceTrack paper. ``DM SORT'' is an abbreviation of Deep MoveSORT. }
        \label{tab:result-dance}
        \end{table}

        \subsubsection{SportsMOT}
        To further demonstrate the performance of MeMoSORT~in fast and variant velocity scenarios, we conduct experiments on the SportsMOT benchmark which is characterized by more strenuous object motion. 
        As detailed in Table \ref{tab:result-sport}, MeMoSORT~establishes a new state-of-the-art with 82.1\% HOTA. 
        It comprehensively surpasses the previous leading method Deep HM-SORT~\cite{gran-henriksen2024deephmsort} across all evaluation criteria. 
        % achieving notable gains of 2.0\% HOTA, 2.9\% AssA, and 1.2\% IDF1.
        It is particularly insightful to compare MeMoSORT~with SportMamba~\cite{khanna2025sportmamba}, which also explores advanced motion modeling but employs a computationally intensive, data-driven Mamba architecture. 
        In stark contrast, our powerful approach not only achieves a remarkable 4.8\% higher HOTA but also demonstrates vastly superior association capabilities. 
        The substantial improvements in AssA and IDF1 (i.e. +8.8\% and +8.7\%) over SportMamba strongly indicate the effectiveness and adaptability of our Mo-IoU, which excels in maintaining target identity against the complex motion characteristic of this benchmark. 
        This result reinforces that a well-designed, method can outperform more complex data-driven counterparts in challenging motion scenarios.

        \begin{table}[h]
        \centering
        \setlength{\tabcolsep}{1mm}
        %\resizebox{.95\columnwidth}{!}{
        \begin{tabular}{l|c c c c c}
            \toprule
            Tracker & HOTA & AssA & IDF1 & DetA & MOTA \\ 
            \midrule
                ByteTrack & 64.1  & 52.3  & 71.4  & 78.5  & 95.9  \\ 
                MambaMOT & 71.3  & 58.6  & 71.1  & 86.7  & 94.9  \\ 
                MambaTrack & 72.6  & 60.3  & 72.8  & 87.6  & 95.3  \\ 
                OC-SORT & 73.7  & 61.5  & 74.0  & 88.5  & 96.5  \\ 
                MotionTrack & 74.0  & 61.7  & 74.0  & 88.8  & 96.6  \\ 
                Mix-SORT & 74.1  & 62.0  & 74.4  & 88.5  & 96.5  \\ 
                ET-Track & 74.3  & 62.1  & 74.5  & 88.8  & 96.8  \\ 
                Track SSM & 74.4  & 62.4  & 74.5  & 88.8  & 96.8  \\ 
                Move SORT & 74.6  & 63.7  & 76.9  & 87.5  & 96.7  \\ 
                EscapeTrack & 75.2  & 65.3  & 80.2  & - & 95.3  \\ 
                DiffMOT & 76.2  & 65.1  & 76.1  & \underline{89.3}  & \textbf{97.1}  \\ 
                Deep-EIoU & 77.2  & 67.7  & 79.8  & 88.2  & 96.3  \\ 
                SportMamba & 77.3  & 66.8  & 77.7  & \textbf{89.5}  & 96.9  \\ 
                DM SORT & 78.7  & 70.3  & 81.7  & 88.1  & 96.5  \\ 
                Deep HM-SORT & \underline{80.1}  & \underline{72.7}  & \underline{85.2}  & 88.3  & 96.6  \\ 
                \textbf{MeMoSORT} & \textbf{82.1}  & \textbf{75.6}  & \textbf{86.4}  & \underline{89.3}  & \underline{97.0}  \\ 
            \bottomrule
        \end{tabular}
        \caption{Performance comparison on the SportsMOT test set. The formatting, abbreviation and evaluation metrics are the same as in Table \ref{tab:result-dance}. Results for ByteTrack and OC-SORT are cited from the original SportsMOT paper.}
        \label{tab:result-sport}
        \end{table}

    \subsection{Ablation Study}
    We conduct ablation studies on the DanceTrack validation set, which concentrate on investigating the impact of different modules, different motion models, different type of IoU on the proposed MeMoSORT.

        \subsubsection{Component Ablation} 
        As shown in Table \ref{tab:abl-component}, we evaluate the contributions of the proposed modules. Evidently, our proposed motion model MeKF has led to significant improvements (i.e., 10.47\% HOTA) compared to the baseline method. This underscores the potential of our methods in capturing the nonlinear motion of objects and accurately estimating their positions. Furthermore, the introduction of the Mo-IoU module results in additional enhancements in metrics related to trajectory consistency, with HOTA improving by 10.13 percentage points.
        
        \begin{table}[h]
        \centering
        %\resizebox{.95\columnwidth}{!}{
        \begin{tabular}{c c c|c c c}
            \toprule
            MeKF & Mo-IoU & ReID & HOTA & FPS \\
            % \textuparrow
            \midrule
                       &            &            & 56.94 & \textbf{74.5} \\
            \checkmark &            &            & 67.41 & 60.8 \\
            \checkmark & \checkmark &            & 77.54 & 49.4 \\
            \checkmark & \checkmark & \checkmark & \textbf{77.91} & 28.8 \\
            \bottomrule
        \end{tabular}
        \caption{Ablation study on the key components of MeMoSORT~using the DanceTrack validation set. `\checkmark' indicates the inclusion of the corresponding component. We report the main performance metric (HOTA) and efficiency (FPS).}
        \label{tab:abl-component}
        \end{table}

        \subsubsection{Different Motion Model} 
        To compare our method with other motion models, we conduct the experiment in Table \ref{tab:abl-motion}, including the linear motin model (KF), data-driven models (LSTM, Transformer, Diffusion), and a baseline without motion modeling. 
        MeKF achieves optimal performance across most metrics with 67.41\% in HOTA,  49.58\% in AssA, 66.41\% in IDF1,  97.55\% in MOTA, 91.69\% in DetA, demonstrating superior robustness in complex motion scenarios. This advantage stems from MeKF's architecture that directly utilizes physical motion priors and data characteristic, yielding more accurate estimation than purely data-driven alternatives.
    
        \begin{table}[h]
        \centering
        \setlength{\tabcolsep}{1mm}
        \begin{tabular}{l|c c c c c}
            \toprule
            ~ & HOTA & AssA & IDF1 & MOTA & DetA \\
            \midrule
            None & 55.24 & 32.90 & 46.07 & 96.15 & 92.82 \\
            KF & 56.94 & 34.92 & 48.18 & 96.35 & 92.91 \\
            LSTM & 60.16 & 38.97 & 52.31 & 96.64 & 92.94 \\ 
            Transformer & 64.12 & 44.20 & 57.60 & 97.04 & \textbf{93.08} \\
            Diffusion & 65.91 & 46.78 & 60.38 & 97.15 & 92.93 \\
            \textbf{MeKF} & \textbf{67.41} & \textbf{49.58} & \textbf{66.41} & \textbf{97.55} & 91.69 \\
            \bottomrule
        \end{tabular}
        \caption{Performance comparison of different filter within the MeMoSORT~framework on the DanceTrack validation set. For all metrics, higher values indicate better performance. The best results among the compared models are shown in \textbf{bold}.}
        \label{tab:abl-motion}
        \end{table}

        \subsubsection{Different Type of IoU} 
        Table \ref{tab:abl-IoU} compares the effectiveness of E-IoU, HM-IoU, HA-EIoU, Mo-IoU metrics against the standard IoU on the DanceTrack validation set, in which we use MeKF as motion models. 
        EIoU enhances performance by expanding the box to accommodate non-linear motion, achieving a HOTA of 70.80\%.
        More substantially, HMIoU, which emphasizes box height, yields a significant leap to 72.70\% HOTA. 
        By combing both of them, HA-EIoU further improved overall tracking performance with HOTA comes to 75.21\%.
        Ultimately, our proposed Mo-IoU achieves the best performance across all metrics, surpassing all existing variants of IoU. 
        Its success stems from adaptively setting proper parameters to control expansion scale and height importance, leading to more robust and accurate tracking.
        % Crucially, these accuracy gains incur minimal FPS reduction, confirming Mo-IoU's suitability for real-time deployment.
        % As a new adaptive association metric, Mo-IoU adaptively adapts to different motion characteristics in real-time systems by setting proper parameters controlling expansion scale and height importance automatically. The quantization strategy ensures these adaptations incur minimal computational overhead, making Mo-IoU particularly suitable for online tracking systems where both accuracy and efficiency are critical.
    
        \begin{table}[h]
        \centering
        \setlength{\tabcolsep}{1mm}
        \begin{tabular}{l|c c c c c c}
            \toprule
             & HOTA & AssA & IDF1 & MOTA & DetA \\
            \midrule
            IoU & 67.41 & 49.58 & 66.41 & 97.55 & 91.69 \\
            EIoU & 70.80 & 54.37 & 70.50 & 97.62 & 92.24 \\
            HMIoU & 72.70 & 57.15 & 71.65 & 97.66 & 92.52 \\
            HA-EIoU & 75.21 & 60.97 & 74.53 & 97.71 & 92.81 \\
            \textbf{Mo-IoU} & \textbf{77.54} & \textbf{64.73} & \textbf{76.92} & \textbf{97.74} & \textbf{92.93} \\
            \bottomrule
        \end{tabular}
        \caption{Performance comparison of different IoU variants within the MeMoSORT~framework on the DanceTrack validation set. The formatting and evaluation metrics are the same as in Table \ref{tab:abl-motion}.}
        \label{tab:abl-IoU}
        \end{table}

    \subsection{Case Analysis}
    
    \begin{figure*}[t]
        \centering
            \subfigure[Case 1: DiffMOT]{
                \label{fig: result-case1a}
                \includegraphics[width=0.45\linewidth]{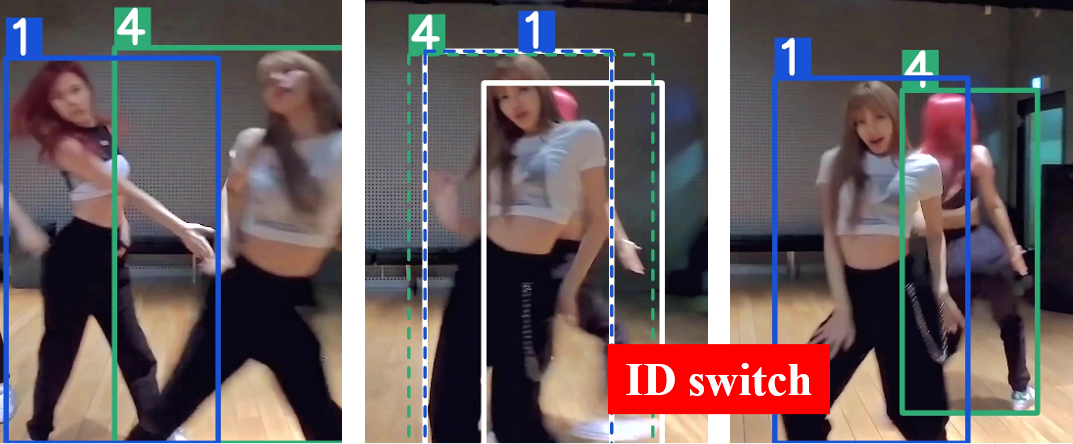}
            }
            \quad
            \subfigure[Case 1: MeMoSORT]{
                \label{fig: result-case1b}
                \includegraphics[width=0.45\linewidth]{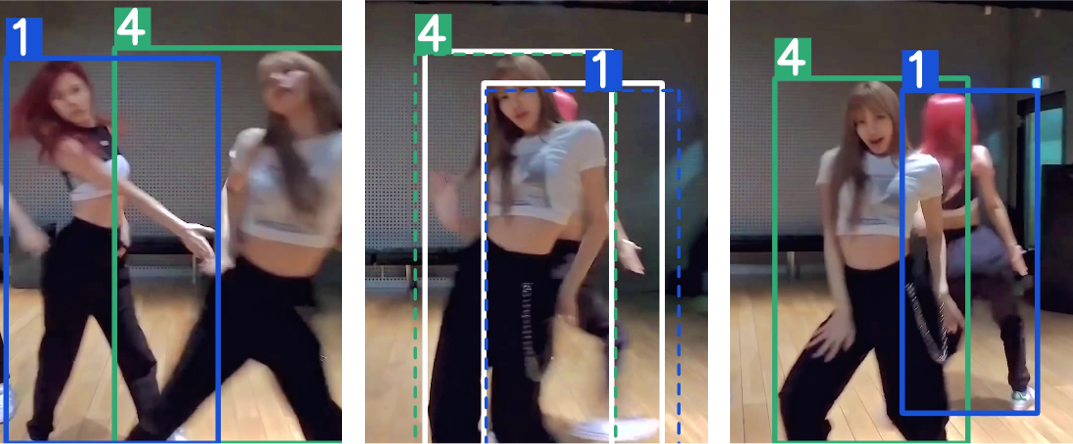}
            }
            \\
            \subfigure[Case 2: DiffMOT]{
                \label{fig: result-case2a}
                \includegraphics[width=0.45\linewidth]{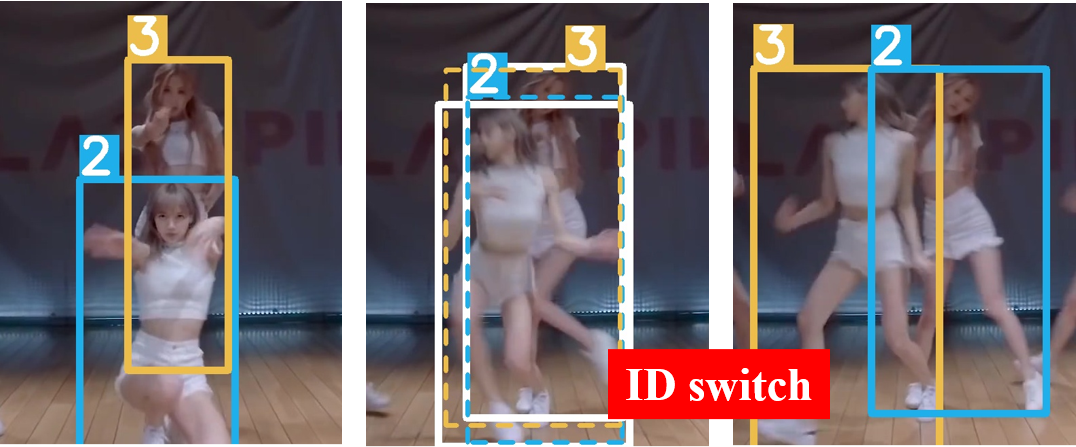}
            }
            \quad
            \subfigure[Case 2: MeMoSORT]{
                \label{fig: result-case2b}
                \includegraphics[width=0.45\linewidth]{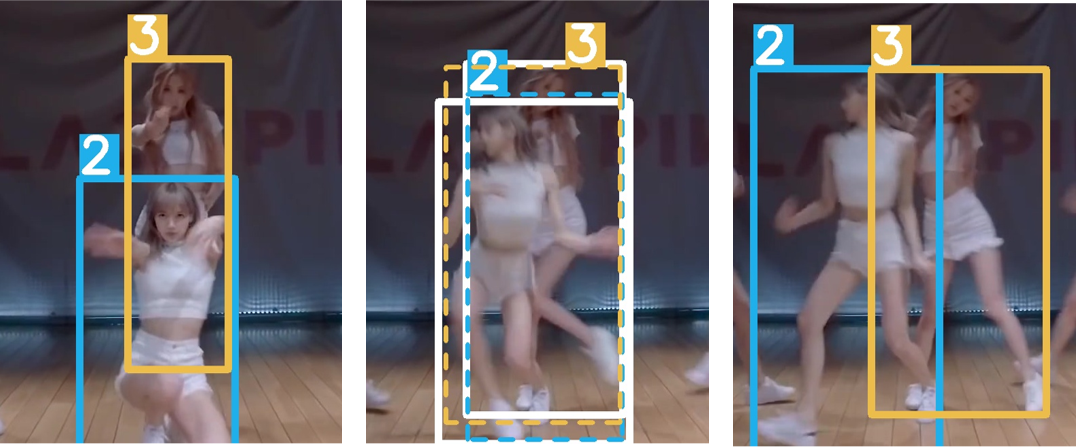}
            }
        \caption{
        Qualitative analysis of DiffMOT (a, c)  and MeMoSORT (b, d) in challenging scenarios from the DanceTrack validation set.
        Solid and dashed colored box represents the trajectory and prediction box, respectively. Solid white box denotes the detection box. 
        \textbf{Case 1 (Complex Motion)}: Both trackers initialized trajectory ID 1 (red hair) and ID 4 (blonde hair). During movement, DiffMOT's inaccurate prediction leads to an identity switch, while MeMoSORT maintains the correct identity by leveraging the precise state estimation from its MeKF.
        \textbf{Case 2 (Severe Occlusion)}: Large positional mismatch between the prediction and the actual detection after occlusion causes the standard IoU-based association in DiffMOT to fail. MeMoSORT's Mo-IoU robustly handles this challenge and ensuring continuous tracking.
        }
        \label{fig: result}
    \end{figure*}

    To intuitively demonstrate the effectiveness of our proposed MeMoSORT, we present case analysis on the challenging sequences from the DanceTrack validation set in Figure \ref{fig: result}. 
    DiffMOT fails in both complex motion (case 1) and severe occlusion (case 2), which both lead to ID switch. In contrast, MeMoSORT, leveraging precise state estimation from the MeKF and the robust Mo-IoU for association, effectively handles both complex motion and severe occlusion, maintaining correct identities and ensuring continuous tracking. 
    This highlights the effectiveness of MeMoSORT's key components in challenging MOT scenarios.

\section{Conclusion}
In this paper, we present MeMoSORT, a simple, online and real-time MOT tracker designed to overcome key limitations in conventional TBD methods. 
Our approach introduces two key innovations: the Memory-assisted Kalman Filter (MeKF), which uses a memory-augmented neural network to correct motion estimation errors, and the Motion-adaptive IoU (Mo-IoU), which adaptively expands the matching space and incorporates height similarity to ensure robust data association.
The effectiveness of our method is demonstrated through extensive experiments, where MeMoSORT achieves SOTA performance on the challenging benchmark DanceTrack and SportsMOT, providing a robust solution for MOT challenges.

\bibliography{aaai2026}

\end{document}